BY NOORA AL-EMADI, MUHAMMAD IMRAN, YIN YANG, INGMAR WEBER, FABJAN LASHI, GAIA RIGODANZA, IVANA HAJŽMANOVÁ, AND FERDA OFLI

# AI-Driven Disaster Response and Displacement Monitoring

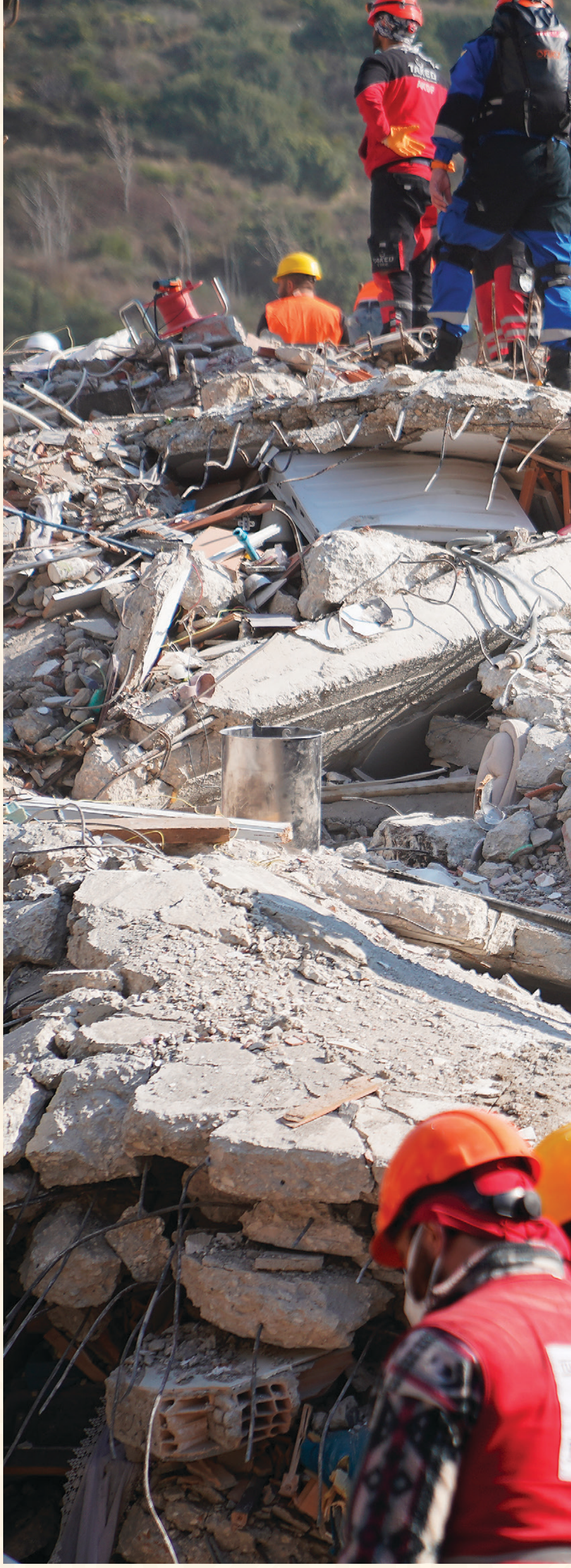

ARTIFICIAL INTELLIGENCE (AI) has demonstrated its potential across a wide range of humanitarian tasks, offering scalable and data-driven solutions to some of the world's most pressing challenges. For instance, in addressing climate change, AI-powered models are being used for faster and more accurate global weather forecasting,[13] control and design of renewable energy,[2] and monitoring deforestation.[19] Machine learning algorithms analyze satellite imagery and socioeconomic data to predict poverty[11] and promote sustainable development.[4] AI has also transformed disaster response and conflict monitoring, providing innovative solutions for assessing damage and urgent needs[9] and tracking displacement.[20]

This article showcases two projects that exemplify the potential of AI-driven approaches for disaster response and displacement monitoring with case studies from the Arab world. The first case study highlights how social media data was utilized to support disaster response efforts during the devastating 2023 Türkiye-Syria earthquake.[24] By processing vast amounts of unstructured social media content, AI-powered tools enabled real-time situational awareness, facilitating more targeted and efficient

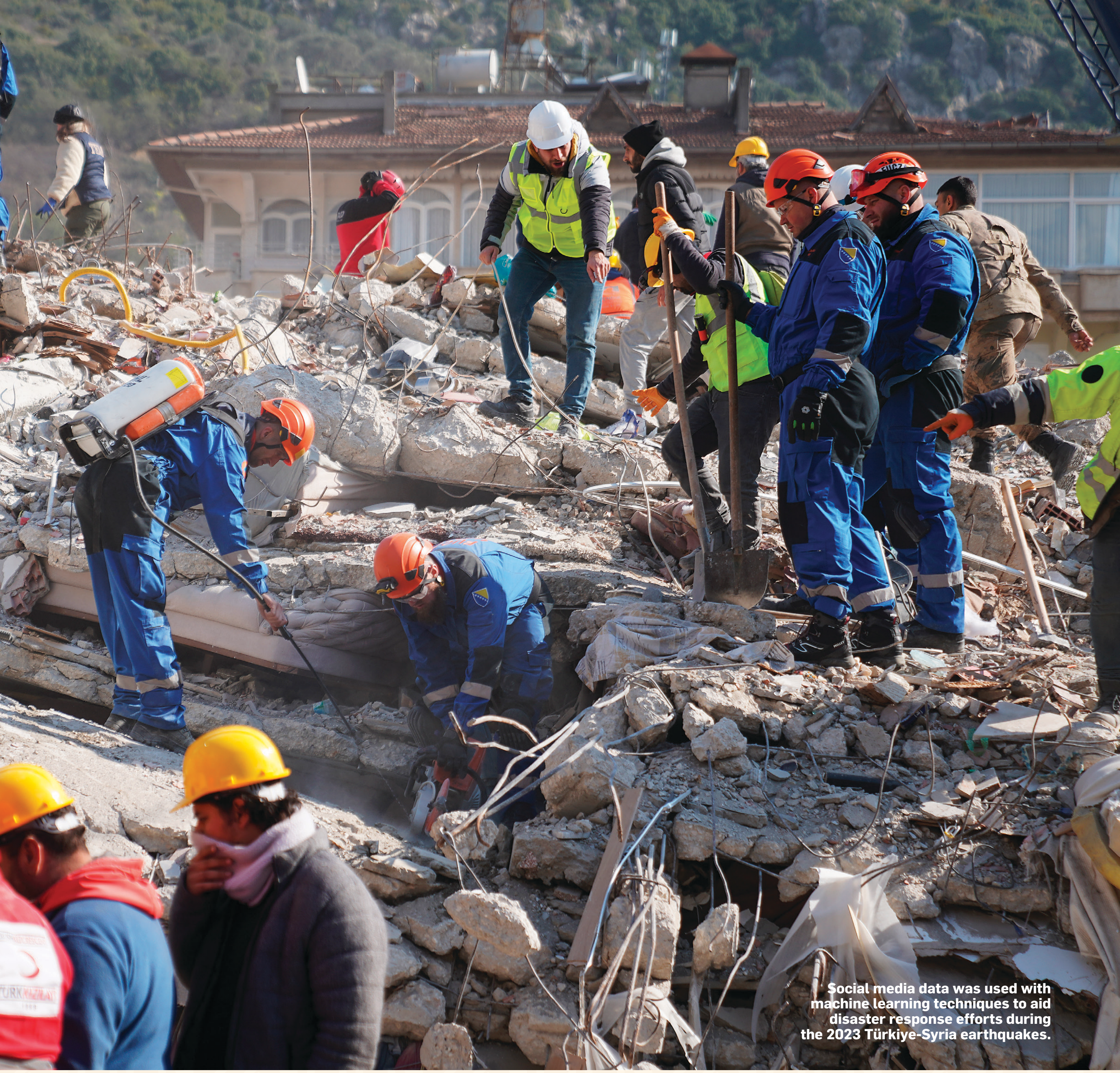

Social media data was used with machine learning techniques to aid disaster response efforts during the 2023 Türkiye-Syria earthquakes.

humanitarian interventions. The second case study focuses on the use of satellite imagery to monitor internal displacement during the Syrian civil war.[23] Through the development of a region-specific vehicle detection model, we demonstrate how vehicle counts in satellite images can serve as proxies for tracking population movements. This innovative approach addresses critical data gaps in conflict-affected regions, offering a scalable solution for monitoring displacement trends in near real time.

## Disaster Response through Social Sensing after the 2023 Türkiye-Syria Earthquakes

**Background.** The 2023 Türkiye-Syria earthquakes, also known as the 2023 Kahramanmaraş earthquakes, were two catastrophic events that struck nine hours apart on February 6, 2023, with epicenters in the Pazarcık and Elbistan districts of Kahramanmaraş, and magnitudes of 7.8 Mw and 7.5 Mw, respectively (see Figure 1).

These earthquakes caused widespread devastation in both Türkiye and Syria, resulting in at least 53,537 deaths in Türkiye and at least 8,476 deaths in Syria, with more than 122,000 people injured in total, according to official figures. However, our study focused on Türkiye due to the limited availability of social media data from Syria. Following the earthquake, widespread destruction across 11 provinces (125 districts) in southern Türkiye created an urgent need for resources and timely information. In response, the Crisis Computing team at the Qatar Computing Research Institute (QCRI), in collaboration with the United Nations Development

**By processing vast amounts of unstructured social media content, AI-powered tools enabled real-time situational awareness, facilitating more targeted and efficient humanitarian interventions.**

Programme (UNDP), leveraged social media data to provide real-time situational awareness. Within a month of the disaster, the team processed over 94 million tweets and 3.6 million images from Twitter (now X), using event-related keywords to identify relevant content. To enhance the utility of this data, tweets mentioning specific locations, such as provinces and districts, were filtered, resulting in approximately two million geotagged tweets and 627,000 images. These geotagged tweets were further analyzed to identify missing people reports, shelter requests or offers, and general help requests, while the associated images were examined for signs of structural damage reports. The temporal distribution of these information categories over the one-month period is illustrated in Figure 2.

**Methodology.** The unstructured nature and high volume of social media data made manual processing impractical. To address this, we applied machine learning techniques, including natural language processing and image recognition, to efficiently filter, categorize, and analyze incoming data. The image processing pipeline focused on assessing image relevance, eliminating duplicates, and categorizing damage reports,[10] while the text processing pipeline aimed to identify missing people reports, shelter requests or offers, and general help requests. The text analysis employed two distinct approaches: keyword-based matching and a multilingual large language model (LLM), Mistral-NeMo,[15] in a Zero-Shot setting where the prompt did not include examples to guide the model's responses. In the LLM-based approach, we experimented with two types of prompts: a concise (short) prompt and a more detailed (long) prompt that incorporated category descriptions. To evaluate performance, we constructed a ground-truth dataset comprising 500 tweets randomly sampled from the two million geotagged tweets. The results of all experiments are summarized in the table here. This can be attributed to the fact that detailed category descriptions narrow the model's focus, often improving precision by reducing ambiguity. However, this specificity can lead to a notable drop in recall, as the model becomes less likely to identify instances that deviate from the given descriptions. In contrast, concise prompt relies more on the model's inherent understanding of the categories, leading to higher recall and, consequently, better overall F1-scores. Despite this, both LLM variations significantly outperformed the keyword-based matching, which demonstrated limited recall capability.

**Figure 1. The 2023 Türkiye-Syria earthquakes. Image credit: U.S. Geological Survey.[22]**

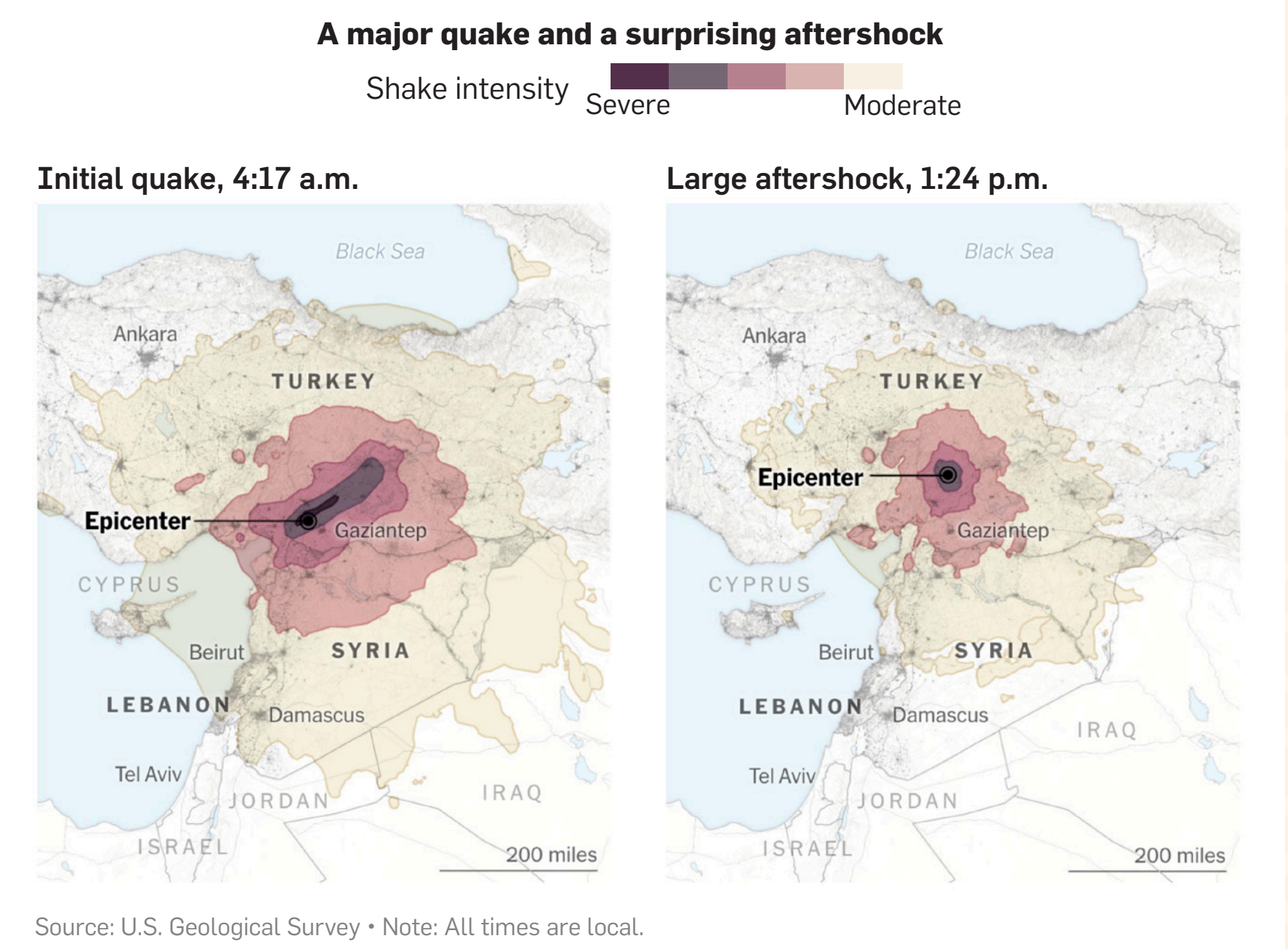

**Figure 2. Categorywise distribution of tweets classified by the Mistral-NeMo model using a concise prompt.**

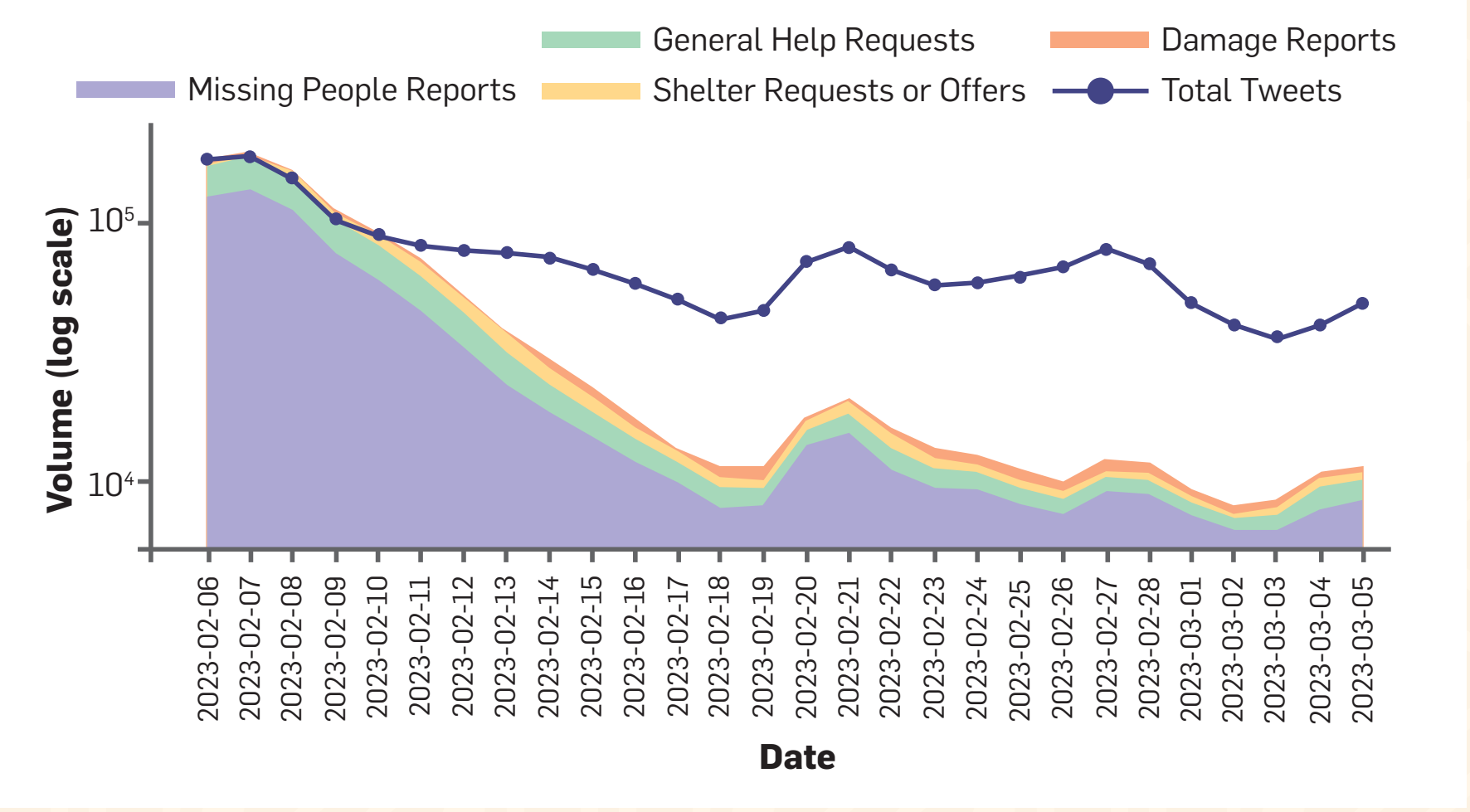

**Table. Categorization performance of the classification models with highest F1-scores underlined.**

| | Keyword Matching | | | LLM (Long Prompt) | | | LLM (Short Prompt) | | |
|---|---|---|---|---|---|---|---|---|---|
| **Category** | **P** | **R** | **F1** | **P** | **R** | **F1** | **P** | **R** | **F1** |
| General Help Requests | 0.49 | 0.24 | **0.32** | 0.69 | 0.23 | **0.34** | 0.55 | 0.48 | **0.51** |
| Missing People Reports | 1.00 | 0.02 | **0.04** | 0.46 | 1.00 | **0.63** | 0.52 | 0.99 | **0.69** |
| Shelter Requests or Offers | 1.00 | 0.20 | **0.33** | 0.75 | 0.38 | **0.50** | 0.73 | 0.40 | **0.52** |
| Micro-Average | 0.58 | 0.13 | **0.21** | 0.50 | 0.63 | **0.56** | 0.54 | 0.72 | **0.62** |

**Building actionable maps.** The data processed by Mistral-NeMo (short prompt) was then aggregated into district-level maps as illustrated in Figure 3, visualizing the intensity and location of needs such as missing people reports, shelter requests or offers, general help requests, and areas impacted by damage. These visual tools supported UNDP and other responders to accelerate their efforts in assessing the extent of specific information, such as the number and location of shelter requests. Additionally, our social sensing data provided insights into the immediate needs of the affected population—information that would typically require conducting a survey.

**Impact and lessons learned.** This interdisciplinary collaboration between response organizations and data scientists highlights the value of AI and social media in disaster response, offering a model for rapid, scalable, and effective humanitarian assistance. While the current results are not perfect, they reveal opportunities for improvement. One potential enhancement is adopting a Few-Shot approach, where examples are included alongside class descriptions to provide the model with additional context.[5,12] A more advanced refinement could involve fine-tuning the model, although this would require a larger volume of labeled data and computational resources. Additionally, information derived from social media posts should be validated against other sources, given the high risk of misinformation and false reporting.

### Monitoring Internal Displacement from Space during the Syrian Civil War

**Motivation.** Access to reliable mobility data is crucial for understanding human movement patterns, especially in conflict-affected regions where such data is often scarce and hard to obtain. To help fill the spatiotemporal gap, vehicle detection and counting in satellite imagery can serve as a valuable proxy for tracking population movements, as shifts in vehicle distributions are tied to shifts in population distributions. This approach holds particular promise in the Middle East, a region marked by high conflict activity and minimal cloud cover, which allows for consistent satellite imaging. However, existing vehicle detection models are rarely trained on data specific to this region, limiting their precision and effectiveness.

**VME model.** To address this, QCRI recently introduced the Vehicles in the Middle East (VME) dataset,[3]

**Figure 3. District-level aggregation of important information processed by Mistral-NeMo using short prompt.**

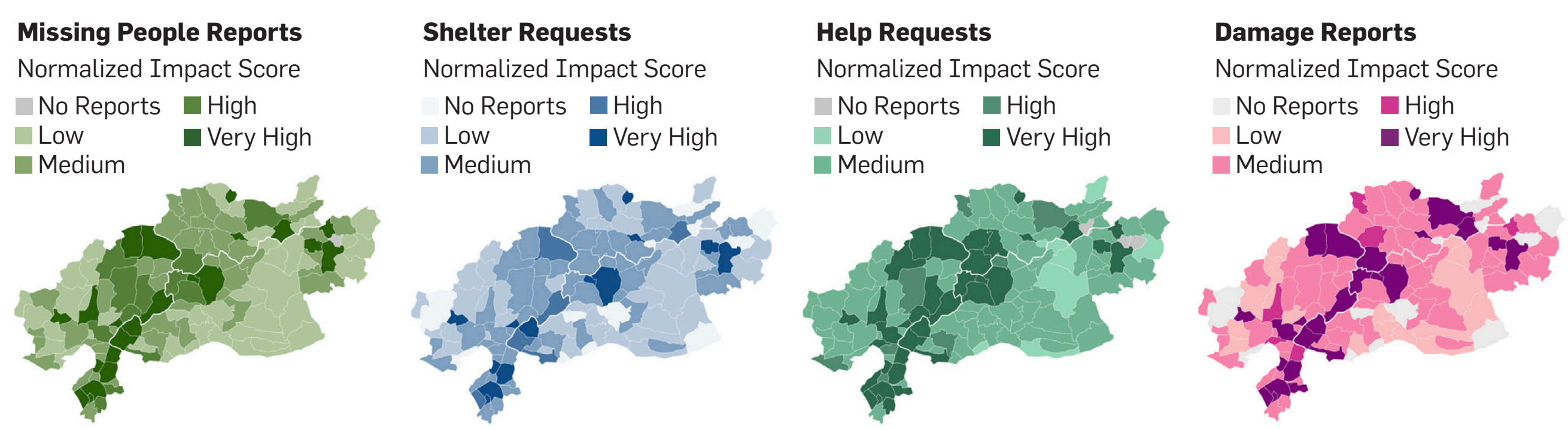

Figure 4. Geographical locations of image collection sites in Syria, with case study areas highlighted in green.

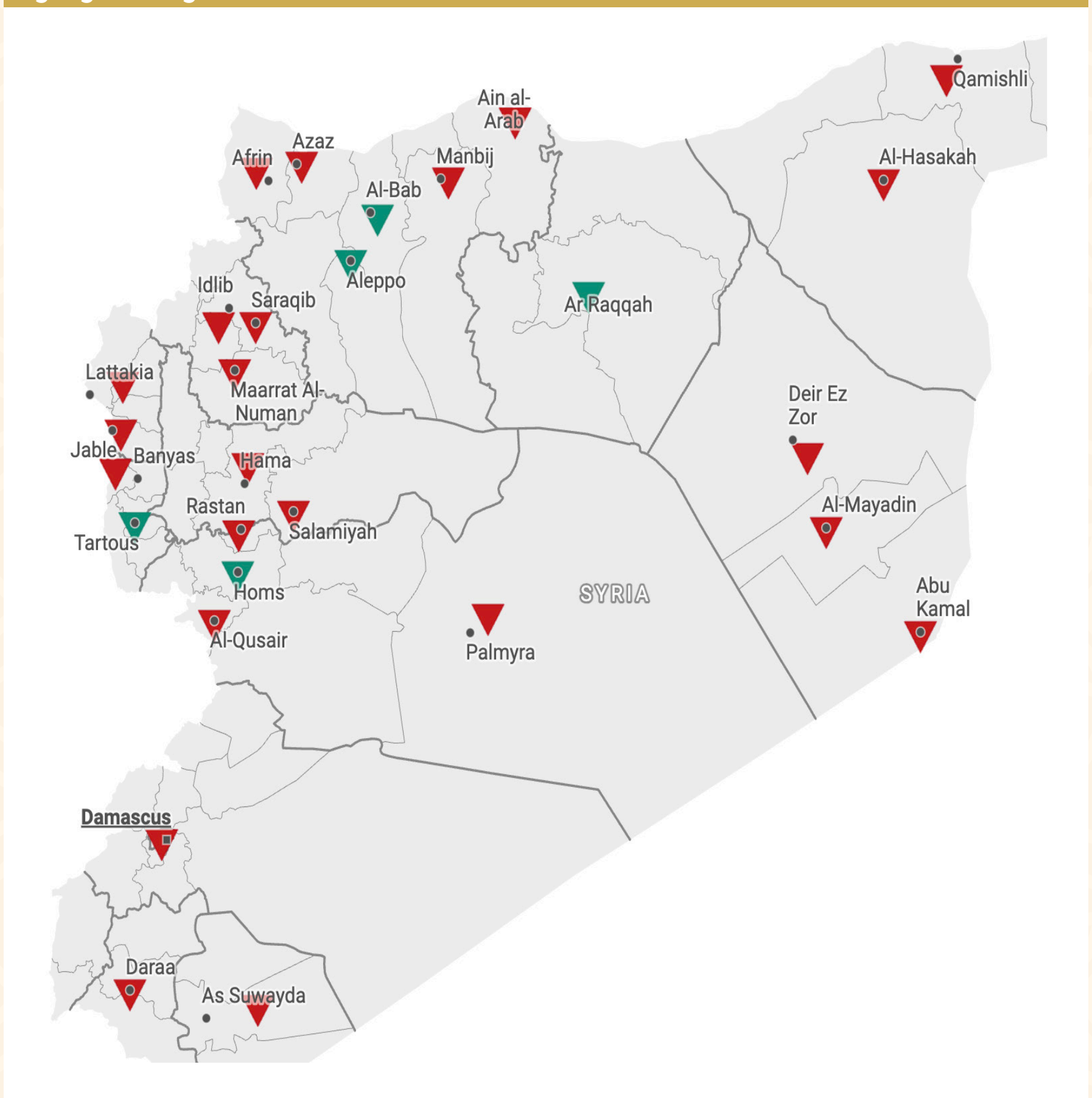

comprising over 4,000 high-resolution satellite image tiles sourced from Maxar Technologies, a satellite imagery provider, and more than 100,000 annotated vehicles across 54 cities in 12 Middle Eastern countries. Our experiments show that the model trained on VME achieves significantly better performance (Precision=0.89 & Recall=0.68) than the baseline (Precision=0.57 & Recall=0.20)[14] in the Middle Eastern context.

**Case study.** This study explores the potential of the VME model to monitor internal displacement during the Syrian civil war, which has displaced approximately 19.6 million people since 2011.[8] Using historical high-resolution satellite imagery from 28 Syrian cities between 2009–2023 (totaling 1,818 images with an average dimension of 22,475 x 24,043 pixels; see Figure 4), we applied our vehicle detection model to track vehicle counts over time and location.

**Findings.** Preliminary analyses reveal plausible parallels between vehicle counts and internal displacement trends when compared with recorded data from sources such as Internal Displacement Monitoring Centre (iDMC), United

Figure 5. The number of cars detected in various cities across Syria during different time periods.

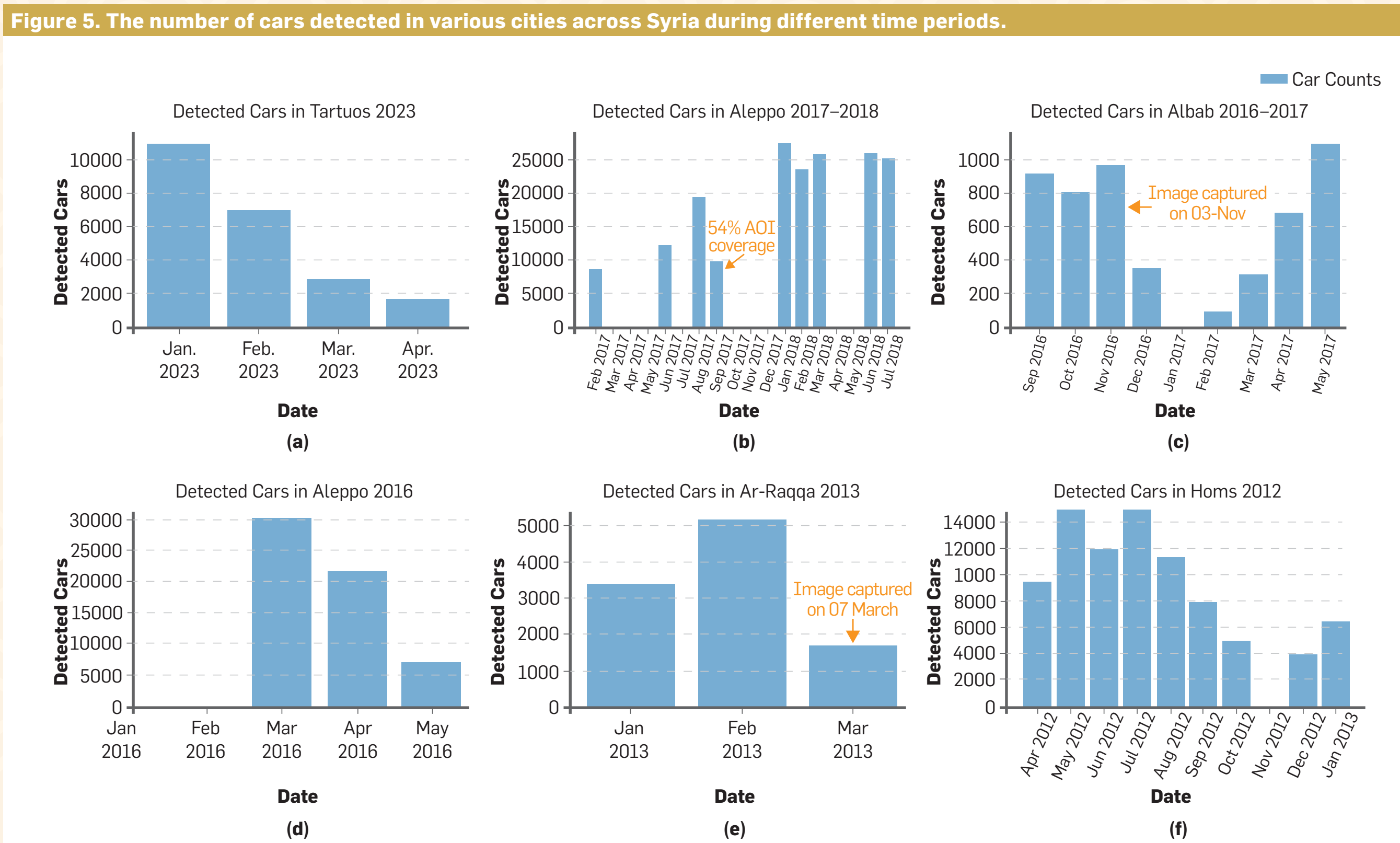

Nations Office for the Coordination of Humanitarian Affairs (OCHA), and Assessment Capacities Project (ACAPS), where these sources may have certain spatial-temporal limitations. According to iDMC, approximately 171,000 people were displaced from Tartous by July 2023. Our analysis of vehicle counts based on available imagery from early 2023 indicates a significant decline in the number of vehicles across the city between January and April (see Figure 5a). Additionally, iDMC highlighted the impact of a local ceasefire in Syria in 2018,[7] which led to the return of families to Idlib and Aleppo. The report estimated that around 66K people were displaced to these two cities. In line with this report, our analysis shows an increase in the number of cars in Aleppo in 2018 compared to 2017; except for a drop in September 2017 due to low area coverage (see Figure 5b). However, no images from Idlib were available in our dataset for this period. Furthermore, a case documented by iDMC and OCHA regarding the Battle of Al-Bab,[6,17] which took place between November 6, 2016, and February 23, 2017, reported that 66K people were displaced. Our analysis indicates a decline in vehicle counts after November 2016 (based on an image captured November 3), followed by an increase after the battle ended in March 2017 (see Figure 5c). In 2016, OCHA documented conflict events and internal displacement between January and May.[16] Our analysis identified a correlation between the reported conflict events in Aleppo during April and May and fluctuations in vehicle counts over the same period (see Figure 5d). In 2013, ACAPS reported[1] two significant displacement events: one in Ar-Raqqa on March 4 and another in Homs by February, causing significant displacement. Our analysis indicates a decline in car counts after these dates, as shown in Figure 5e, with the image captured on March 7 and in Figure 5f, presenting the data from August 2012 to January 2013. These findings suggest that vehicle detection in satellite imagery can effectively support humanitarian organizations in monitoring population displacement in near real time, offering an innovative solution where conventional methods are constrained.

**Limitations and lessons learned.** Satellite imagery is a useful tool for tracking internal displacement, revealing movement patterns that traditional data often misses. While car mobility can indicate displacement in some contexts, such as the Ukraine-Russia war, it is less relevant in regions where people lack access to vehicles, such as Venezuela's Caminantes[18] and Africa.[21] Challenges like inconsistent imagery, limited coverage, and cloud obstruction affected analysis. To fill these gaps, we analyzed pre- and post-event movement patterns. Despite limitations in spatial and temporal resolution, our results show a plausible link between vehicle counts and displacement. Incorporating advanced spatiotemporal modeling could enhance these insights further.

## Conclusion

Together, these case studies illustrate the transformative potential of AI in leveraging non-traditional data sources for humanitarian applications. By integrating cutting-edge computational methods with domain expertise, these approaches pave the way for more informed and impactful decision making in disaster response and conflict monitoring. As we discuss the methodologies, challenges, and implications of these initiatives, this article aims to inspire further research and collaboration at the intersection of AI, data science, and humanitarian action. **C**

**Noora Al-Emadi** is a senior software engineer at Qatar Computing Research Institute, Hamad Bin Khalifa University, Qatar.

**Muhammed Imran** is a principal scientist at Qatar Computing Research Institute, Hamad Bin Khalifa University, Qatar.

**Yin Yang** is a professor at the College of Science and Engineering, Hamad Bin Khalifa University, Qatar.

**Ingmar Weber** is an Alexander von Humboldt Professor in AI at Saarland University, Germany.

**Fabjan Lashi** is a digital assessments project manager with the United Nations Development Programme, Austria.

**Gaia Rigodanza** is a research development analyst with the United Nations Development Programme, Austria.

**Ivana Hajžmanova** is a global monitoring manager at the Internal Displacement Monitoring Centre, Switzerland.

**Ferda Ofli** (fofli@hbku.edu.qa) is a principal scientist at Qatar Computing Research Institute, Hamad Bin Khalifa University, Qatar.

**Fabjan Lashi**, **Gaia Rigodanza**, and **Ivana Hajzmanova** are stakeholders from the United Nations and a non-governmental organization that operate in and provide humanitarian data related to the Arab world. **Noora Al-Emadi** was supported by Qatar Research, Development and Innovation Council, GSRA7-1-0421-20022. **Ingmar Weber** was supported by Alexander von Humboldt Professur für KI.